\documentclass[runningheads]{llncs}
\usepackage[T1]{fontenc}
\usepackage{graphicx,verbatim}
\usepackage{amsmath,amssymb,amsfonts}
\usepackage{orcidlink}
\usepackage{graphicx}
\usepackage{textcomp}
\def\BibTeX{{\rm B\kern-.05em{\sc i\kern-.025em b}\kern-.08em
    T\kern-.1667em\lower.7ex\hbox{E}\kern-.125emX}}

\usepackage{xargs}                      
\usepackage[pdftex,dvipsnames]{xcolor}  
\usepackage[colorinlistoftodos,prependcaption,textsize=small]{todonotes}
\newcommandx{\unsure}[2][1=]{\todo[linecolor=red,backgroundcolor=red!25,bordercolor=red,#1]{#2}}
\newcommandx{\change}[2][1=]{\todo[linecolor=blue,backgroundcolor=blue!25,bordercolor=blue,#1]{#2}}
\newcommandx{\info}[2][1=]{\todo[linecolor=OliveGreen,backgroundcolor=OliveGreen!25,bordercolor=OliveGreen,#1]{#2}}
\newcommandx{\improvement}[2][1=]{\todo[linecolor=Plum,backgroundcolor=Plum!25,bordercolor=Plum,#1]{#2}}
\newcommandx{\thiswillnotshow}[2][1=]{\todo[disable,#1]{#2}}

\usepackage{hyperref}
\hypersetup{
    colorlinks=true,
    linkcolor=blue, 
    filecolor=blue,      
    urlcolor=blue,
    citecolor=blue, 
}

\usepackage{multirow}
\usepackage{tabularx}
\usepackage{hhline}
\usepackage{array}
\newcolumntype{P}[1]{>{\centering\arraybackslash}p{#1}}
\newcolumntype{L}[1]{>{\raggedright\let\newline\\\arraybackslash\hspace{0pt}}m{#1}}
\newcolumntype{C}[1]{>{\centering\let\newline\\\arraybackslash\hspace{0pt}}m{#1}}
\newcolumntype{R}[1]{>{\raggedleft\let\newline\\\arraybackslash\hspace{0pt}}m{#1}}
\usepackage{booktabs}

\usepackage{paralist}

\usepackage[normalem]{ulem}
\usepackage{xfrac}

\usepackage[symbol]{footmisc}

\RequirePackage[normalem]{ulem} 
\RequirePackage{xcolor}
\usepackage{soul}
\usepackage{ulem}
\usepackage{censor}
\newlength\nextcharwidth
\makeatletter
\renewcommand\@cenword[1]{%
  \setlength{\nextcharwidth}{\widthof{#1}}%
  \censorrule{\nextcharwidth}%
  \kern -\nextcharwidth%
  #1}
\makeatother

\usepackage{algorithmicx}
\usepackage[ruled]{algorithm}
\usepackage{algpseudocode}

\graphicspath{{figures/}}
\DeclareGraphicsExtensions{.pdf,.jpeg,.png,.eps}
\begin{document}
\title{DenseTRF: Texture-Aware Unsupervised Representation Adaptation for Surgical Scene Dense Prediction}
\titlerunning{DenseTRF: Texture-Aware Unsupervised Representation Adaptation}
%
\author{
Guiqiu Liao\inst{1,2}\orcidlink{0000-0003-0921-178X} \and
Matja\v{z} Jogan\inst{1,2}\orcidlink{0000-0003-3771-3146} \and
Daniel A. Hashimoto\inst{1,2,3}\orcidlink{0000-0003-4725-3104} 
}
\authorrunning{G. Liao et al.}
%
\institute{
GRASP Laboratory, University of Pennsylvania\and
PCASO Laboratory, Department of Surgery,
University of Pennsylvania \and
Department of Computer and Information Science,
University of Pennsylvania
\email{Guiqiu.Liao@pennmedicine.upenn.edu} }


  
\maketitle              
\begin{abstract}
Dense prediction tasks in surgical computer vision, such as segmentation and surgical zone prediction, can provide valuable guidance for laparoscopic and robotic surgery. However, these models often suffer from distribution shifts, as training datasets rarely cover the variability encountered during deployment, leading to poor generalization. We propose DenseTRF, a self-supervised representation adaptation framework based on texture-centric attention. Our method leverages slot attention to learn texture-aware representations that capture invariant visual structures. By adapting these representations to the target distribution without supervision, DenseTRF significantly improves robustness to domain shifts. The framework is implemented through conditioning dense prediction on slot attention and model merging strategies. Experiments across multiple surgical procedures demonstrate improved cross-distribution generalization in comparison to state-of-the-art segmentation models and test-distribution adaptation methods for dense prediction tasks. {Code shared at: \href{https://github.com/PCASOlab/Dense-TRF}{https://github.com/PCASOlab/Dense-TRF}.}
\setcounter{footnote}{0}
\footnotetext{Accepted to 29th International Conference on Medical Image Computing
and Computer Assisted Intervention (MICCAI 2026)}

\end{abstract}
\section{Introduction} 
\label{sec:intro}
Dense prediction tasks in surgical video analysis, such as tissue segmentation, instrument tracking, and surgical zone prediction, are critical for computer-assisted interventions and real-time intraoperative guidance \cite{twinanda2016endonet,maier2024metrics}. Nevertheless, robust deployment remains difficult due to pronounced domain shifts across operating environments caused by variations in anatomy, appearance, viewpoint, illumination, and procedural phase \cite{madani2022artificial,hashimoto2018artificial}.
 
Contemporary surgical scene understanding approaches predominantly rely on supervised learning with densely annotated datasets \cite{ronneberger2015unet,cheng2022mask2former,twinanda2016endonet}. Although these methods achieve strong performance when training and test distributions are aligned, their accuracy degrades under domain shift \cite{hashimoto2018artificial}. This limitation is particularly pronounced in surgery, where dense annotation is costly and labor-intensive, often restricting training data to a limited number of procedures and institutions \cite{janowczyk2016deep,maier2024metrics}. Consequently, models that perform well on development datasets often fail to generalize across clinical settings and procedure types.

Recent advances in foundation vision models \cite{caron2021dino,kirillov2023sam,simeoni2025dinov3} have demonstrated remarkable generalization capabilities by leveraging large-scale pretraining. However, as we show in this work, even these powerful representations benefit from targeted adaptation when deployed on specialized surgical domains with limited annotated data. Test-distribution adaptation methods \cite{prabhudesai2023test,nguyen2024adapting,wang2021tent,Wang_2022_CVPR,xie2022unsupervised} offer a promising direction by adjusting model representations using only unlabeled target data, yet existing approaches often struggle with the texture-rich yet geometrically variable nature of surgical scenes.
 
A complementary line of research has explored object-centric representations through slot-based attention mechanisms \cite{locatello2020object,seitzer2022bridging}. Recent work has further extended slot learning to foundation-model feature spaces, enabling unsupervised object discovery using pretrained representations such as DINO \cite{seitzer2022bridging,didolkar2024zero}. These advances demonstrate that slot attention can capture semantically meaningful structures without supervision. Concurrently, adaptation-oriented extensions have begun to emerge, including slot-based test-time adaptation approaches \cite{prabhudesai2023test} and cross-domain slot learning frameworks \cite{liao2025forla}. Despite these developments, the role of object-centric representations for unsupervised test-distribution adaptation in surgical dense prediction remains largely unexplored. Notably, representations that group visual features by appearance rather than spatial configuration may be particularly advantageous in surgical scenes, where tissue textures exhibit consistent statistical properties even as shapes deform across patients and procedural stages.

We introduce DenseTRF (Dense Texture-Centric Representation Adaptation Framework), an object-centric approach for unsupervised test-distribution adaptation in surgical dense prediction. Our key insight is that slot attention can be steered toward texture-aware representations that capture invariant visual structures across domains. DenseTRF integrates three components: (i) conditioning dense prediction on unsupervised slot representations learned via reconstruction, (ii) a periodic model-merging strategy that balances target specialization with generalization, and (iii) evaluation across three surgical datasets (Thoracic, POEM and Cholec) under extreme low-data regimes (1–5\% annotations). Results demonstrate consistent improvements over state-of-the-art segmentation and foundation-model baselines. Through comprehensive ablation studies, we demonstrate the contribution of each component and show that our representation adaptation strategy and object-centric representations work in synergy, unlocking flexible and scalable applications of dense prediction in surgical domains.

\section{Method}

\begin{figure}[t!]
    \centerline{
    \includegraphics[width=1.0\linewidth]{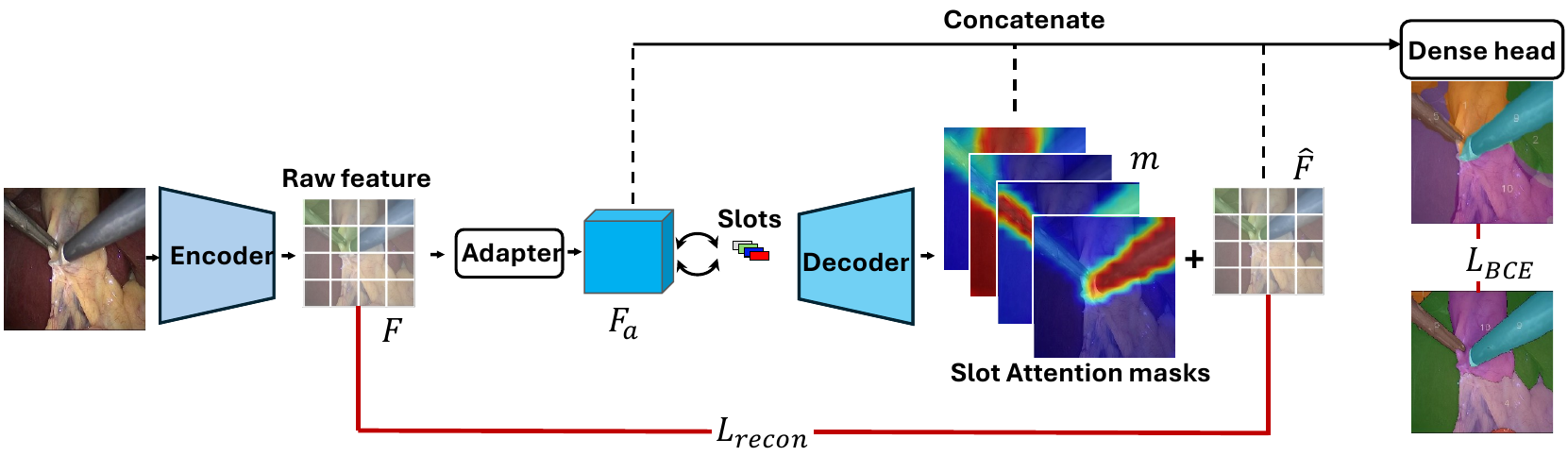}}
    \caption{Architecture of the proposed network, where the dense head is conditioned on object-centric representations learned via Slot Attention and optimized using both reconstruction and supervised dense prediction objectives. } 
    \label{fig_method1}
\end{figure}

\subsection{Network architecture}
Our dense prediction network is built upon object-centric representations learned via Slot Attention (SA). Given an input image, we extract features using a pretrained foundation model encoder~\cite{simeoni2025dinov3}, yielding $F \in \mathbb{R}^{H \times W \times C_r}$. A lightweight MLP adapter $g$ projects these into adapted features $F_a = g(F) \in \mathbb{R}^{H \times W \times C_a}$, which are passed to a SA encoder~\cite{locatello2020object} that iteratively refines $K$ slot latent vectors $\{s_k\}_{k=1}^K$. Following DINOSAUR~\cite{seitzer2022bridging}, MLP decoders reconstruct the original features from the slots, producing per-slot feature maps $\hat{F}_k$ and alpha masks $\alpha_k$:
\begin{equation}
\hat{F} = \sum_{k=1}^K \hat{F}_k \odot m_k, \quad m_k = \text{softmax}_k(\alpha_k), \quad \mathcal{L}_{\text{recon}} = \|F - \hat{F}\|^2
\end{equation}
This learning objective encourages the slots to capture coherent object-like regions, producing interpretable attention masks as a byproduct.
We initially considered using only the adapted features \(F_a\) as input to the dense prediction head. However, we observed that this underutilized the rich object-centric structure captured by the slots. Therefore, we concatenate three sources of information: (i) the adapter output features, (ii) the reconstructed features, and (iii) the SA masks \(m_k\), forming a combined representation per spatial location:
\begin{equation}
z_{ij} = \left[\, f_{ij}^{\text{ada}},\; \hat{f}_{ij}^{\text{recon}},\; m_{1,ij}, \dots, m_{K,ij} \,\right]
\end{equation}
This is fed into a lightweight MLP classifier whose logits are upsampled to the input resolution. The model is trained jointly with:
\begin{equation}
\mathcal{L} = \mathcal{L}_{\text{BCE}} + \lambda\,\mathcal{L}_{\text{recon}}
\end{equation}

DenseTRF is motivated by the need to combine generalizable foundation features with target-specific adaptation under surgical domain shift. The adapter provides a lightweight task-specific projection, while Slot Attention encourages structured grouping of coherent appearance patterns.
\begin{figure}[t!]
    \centerline{
    \includegraphics[width=1.0\linewidth]{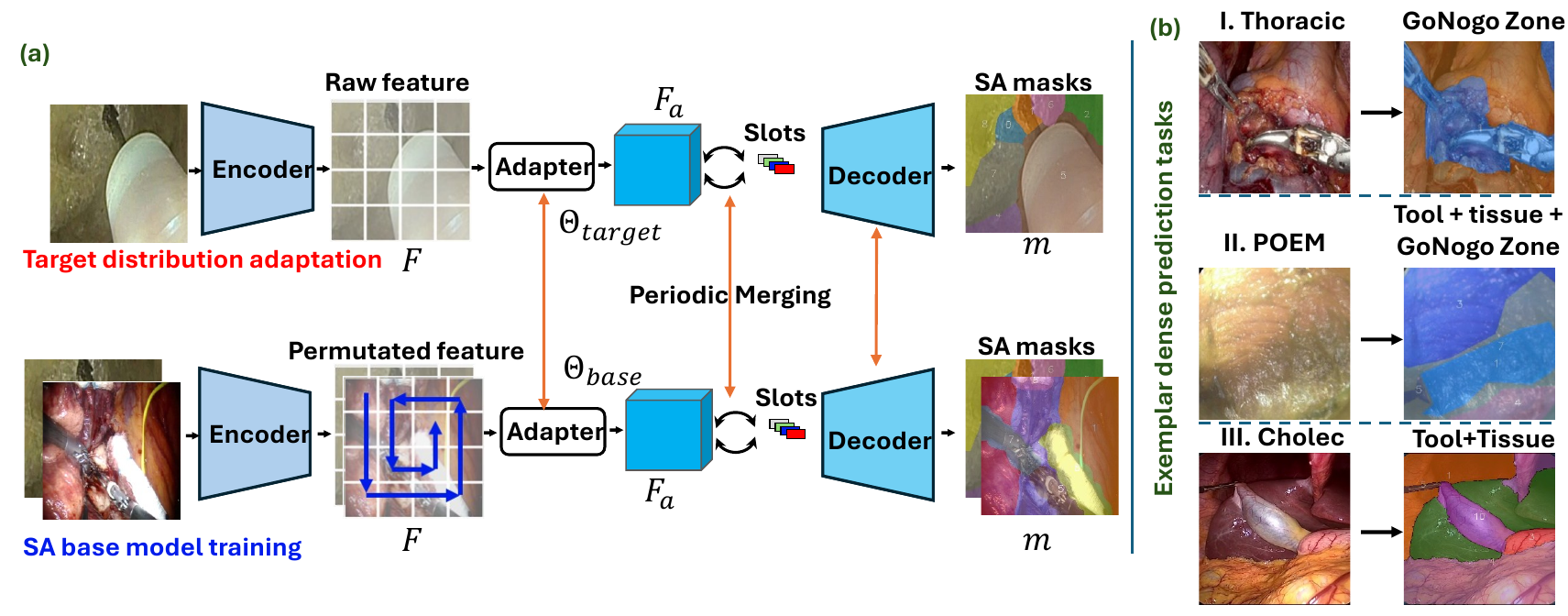}}
    \caption{(a) Slot Attention (SA) adaptation via periodic model merging, which anchors the adapted model to the base representation during specialization. (b) Example dense prediction tasks across Thoracic, POEM, and Cholec surgical datasets.} 
    \label{fig_method2}
\end{figure}

\subsection{Unsupervised test-distribution adaptation}
 
When SA is trained on a mixture of diverse visual domains, the learned slots must explain substantial appearance variability, which can reduce their ability to specialize to a specific deployment scenario. This is especially relevant in surgical settings, where a model's success will be measured by its performance within a particular surgical case and procedure, with video information that exhibits relatively consistent visual statistics compared to all possible surgeries.
 
To address this challenge, we propose an unsupervised test-distribution adaptation strategy that enables SA to retain broad generalization while refining its representation for the target test domain. Our key observation is that when SA is trained jointly on data of any given domains, its attention masks learn to optimize reconstruction loss of the images of training domains. This has an advantage over training exclusively on a single domain, but will lead to poor generalization due to representation drift and bias toward a narrow set of concepts~\cite{liao2025forla}. 

To balance specialization and generalization, we adopt a periodic model merging strategy inspired by continual learning approaches~\cite{yang2025continual}. As shown in Figure \ref{fig_method2}, specifically, we maintain two SA branches that share the same architecture: 

\begin{itemize}
    \item \textbf{Base branch:} trained on a broad multi-domain image pool.
    \item \textbf{Target branch:} trained solely on unlabeled target-domain images.  
\end{itemize}
After each adaptation round, the parameters of the two branches are merged through weighted averaging:
\begin{equation}
\Theta^{(r+1)} = \frac{1}{2}  ( \Theta_{base}^{(r)} + \Theta_{target}^{(r)})
\end{equation}
where $\Theta$ denotes the parameters of the adapter, iterative competitive SA module, and slot decoder. The merged model is then broadcast back to both branches to initialize the next round.

This periodic synchronization serves as an anchoring mechanism that prevents target-domain over-specialization while still allowing progressive refinement of slot representations toward the deployment distribution.

In addition, we design the SA module to learn texture-aware representations that capture invariant visual structures rather than object geometry. We adopt patch order permutation from SPOT~\cite{kakogeorgiou2024spot} to weaken spatial bias and encourage texture-centric slot grouping. This relaxation promotes more robustness when tissue shapes deform or when the camera viewpoint changes, and decouples slot position from appearance.

\section{Experiments and Results} 
\label{sec:experiments_results}

\subsection{Datasets and Metrics}
We evaluate our approach on three surgical video datasets.  (1) \textbf{Thoracic}: 30 sequences (2K unlabeled frames each) collected from robot-assisted lung malignancy performed at the Toronto General Hospital between 2014 and 2023, with 688 annotated go/no-go zones for training and 169 for test. (2) \textbf{POEM}: 40K total Peroral endoscopic myotomy (POEM) images curated from Hospital of the University of Pennsylvania and Massachusetts General Hospital, 12K training pool, with 120 test images containing sparse go/no-go zone annotations. (3) \textbf{Cholec80} \cite{twinanda2016endonet} (24K frames) for unsupervised pretraining and \textbf{CholecSeg8K} \cite{hong2020cholecseg8k} (8K annotated frames): 7K training pool (avoiding temporal overlap), 1k for testing. For all datasets, we use only 1\%–5\% of training pool labels. We report Intersection over Union (IoU), DICE, and Hausdorff Distance (HD in pixels) \cite{maier2024metrics} for region overlap and boundary accuracy.

\subsection{Training Details}

\noindent\textbf{Slot Attention pretraining.} Base branch: LR $4\times10^{-4}$, weight decay $10^{-5}$, with SPOT patch-order permutation and alternating batch schedule.

\noindent\textbf{Test-distribution adaptation.} Target branch: LR $1\times10^{-4}$, weight decay $10^{-5}$, trained only on unlabeled target images with reconstruction loss (no feature order permutation).

\noindent\textbf{Dense prediction head.} LR $1\times10^{-4}$, no weight decay. Loss: BCE + $\lambda \mathcal{L}_{\text{recon}}$ ($\lambda=0.1$). Three-phase training: (i) head only (1k iter), (ii) full model joint, (iii) remove $\mathcal{L}_{\text{recon}}$ (final 1k iter). Total 5k iterations. Early stopping monitored but was never triggered.

All methods, including the proposed approach and competing state-of-the-art baselines, are trained independently across five runs with different random seeds. DenseTRF remains lightweight in practice: both training and inference can be performed with less than 2 GB of GPU memory.


\subsection{Comparison to state-of-the-art segmentation architectures }

\begin{table}[t!]
\caption{Comparison of mDICE scores and model parameter sizes across different methods.}
\label{tab_mdice}
\centering
\setlength\arrayrulewidth{0.9pt}
\setlength\doublerulesep{0.9pt} 
\setlength{\tabcolsep}{5pt} 
\resizebox{0.95\linewidth}{!}{%
\scriptsize 
\begin{tabular}{l |c| c c c}
\hline
Model & Para [MB] & Thoracic & POEM & Cholec \\
\hline
Mask2former\cite{cheng2022mask2former}      & 421 & $68.14 \pm 0.89$ & $32.55 \pm 1.17$ & $36.73 \pm 3.35$ \\
Segman\cite{gao2024segman}          & 202 & $55.52 \pm 4.46$ & $27.99 \pm 1.07$ & $43.70 \pm 1.75$ \\
Swin-UNETR\cite{hatamizadeh2022swinunetr}      &  88 & $68.18 \pm 0.43$ & $31.70 \pm 0.32$ & $48.83 \pm 1.32$ \\
Unet\cite{ronneberger2015unet} + Resnet\cite{he2016resnet}    & 127 & $71.90 \pm 0.70$ & $41.55 \pm 0.74$ & $52.79 \pm 3.12$ \\
Unet\cite{ronneberger2015unet} + MIT\cite{xie2021segformer}      & 251 & $76.50 \pm 0.17$ & $46.37 \pm 0.43$ & $59.56 \pm 2.19$ \\
DenseTRF (ours)  & 388 & $\mathbf{81.35 \pm 0.33}$ & $\mathbf{51.00 \pm 0.79}$ & $\mathbf{67.40 \pm 1.00}$ \\
\hline
\end{tabular}
}
\end{table}

\begin{figure}[t!]
    \centerline{
    \includegraphics[width=1.0\linewidth]{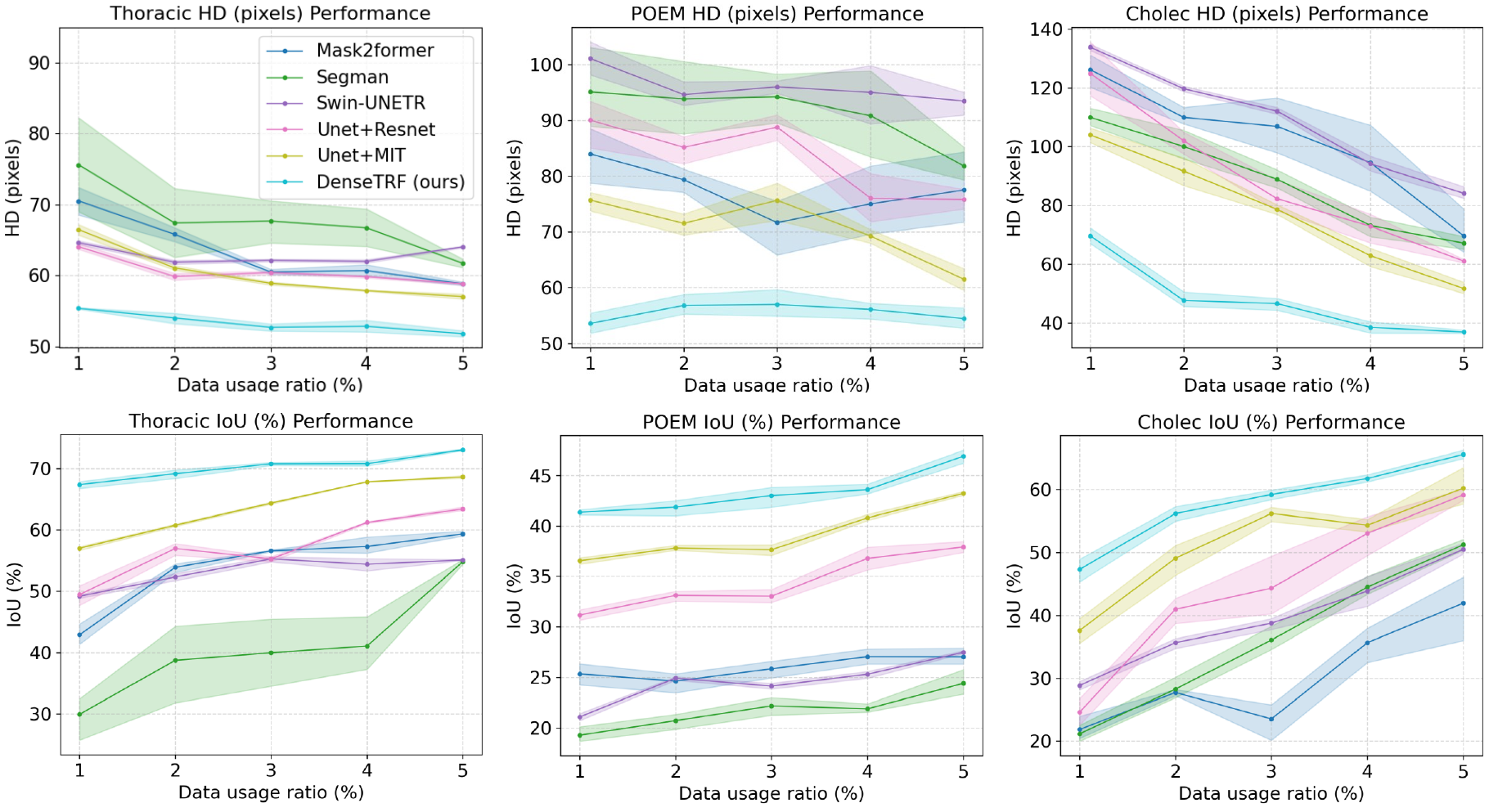}}
    \caption{Comparison with state-of-the-art methods under varying training data ratios using IoU (\%) and Hausdorff Distance (HD, pixels) metrics.} 
    \label{fig_iou_hd}
\end{figure}
We compare DenseTRF against five baseline methods including Mask2Former \cite{cheng2022mask2former}, SegMan \cite{gao2024segman}, Swin-UNETR \cite{hatamizadeh2022swinunetr}, UNet \cite{ronneberger2015unet}+ResNet~\cite{he2016resnet}, and UNet+MIT~\cite{xie2021segformer} across three surgical datasets (Thoracic, Per Oral Endoscopic Myotomy [POEM], and Cholec80) under increasing data-usage ratios from 1\% to 5\%, evaluating both IoU (\%) and Hausdorff Distance (HD, pixels) and DICE score. As shown in Figure \ref{fig_iou_hd}, across all datasets and supervision levels, DenseTRF consistently demonstrates superior dense prediction accuracy, achieving higher IoU scores while simultaneously reducing boundary error as reflected by lower HD values. The performance gap is particularly pronounced in the low-data regime (1–2\%), where DenseTRF maintains stable improvements over competing architectures, indicating enhanced data efficiency and robustness under limited annotation availability.

The mean DICE score across datasets is presented in Table~\ref{tab_mdice}, where DenseTRF achieves the highest performance on Thoracic, POEM, and Cholec80 with mDICE values of $81.35 \pm 0.33$, $51.00 \pm 0.79$, and $67.40 \pm 1.00$, respectively. In comparison to the best among SOTA, UNet+MIT, we improve by $4.85$ mean DICE for Thoracic ($p < 0.05/5$, with Bonferroni correction $m=5$), $4.63$ for POEM ($p < 0.01$), and $7.84$ for Cholec ($p < 0.01$). Collectively, these results demonstrate that DenseTRF provides consistent gains across diverse surgical domains, highlighting its effectiveness in improving dense prediction performance under limited ground truth supervision signal access.

\subsection{Compare to foundation model based adaptation}

\begin{table}[t!]
\caption{Comparison with foundation model-based adaptation under low-data regime.}
\label{tab_FM}
\centering
\setlength\arrayrulewidth{0.9pt}
\setlength\doublerulesep{0.9pt} 
\setlength{\tabcolsep}{5pt} 
\resizebox{1.0\linewidth}{!}{%
\scriptsize 
\begin{tabular}{l| l |c |c c c}
\hline
Dataset & Model & Para [MB] & DICE $\uparrow$ & IoU $\uparrow$ & HD $\downarrow$ \\
\hline
\multirow{6}{*}{Thoracic} 
 & SAM\cite{kirillov2023sam}         & 415 & $73.57 \pm 0.47$ & $59.89 \pm 0.48$ & $60.06 \pm 0.16$ \\
 & DINO-v1\cite{caron2021dino}      & 346 & $74.77 \pm 0.12$ & $61.53 \pm 0.14$ & $58.90 \pm 0.16$ \\
 & CLIP\cite{radford2021clip}         & 393 & $77.76 \pm 0.32$ & $65.16 \pm 0.43$ & $57.76 \pm 1.37$ \\
 & DINO-v3\cite{simeoni2025dinov3}      & 342 & $77.66 \pm 0.39$ & $65.48 \pm 0.47$ & $56.98 \pm 0.46$ \\
 & Slot-TTA\cite{prabhudesai2023test}     & 384 & $73.20 \pm 0.54$ & $59.99 \pm 0.57$ & $70.73 \pm 1.00$ \\
 & \textbf{DenseTRF (ours)} & {388} & $\mathbf{79.32 \pm 0.48}$ & $\mathbf{67.35 \pm 0.58}$ & $\mathbf{55.35 \pm 0.09}$ \\
\hline
\multirow{6}{*}{POEM}
 & SAM\cite{kirillov2023sam}          & - & $38.38 \pm 0.35$ & $31.10 \pm 0.30$ & $78.12 \pm 2.53$ \\
 & DINO-v1\cite{caron2021dino}      & - & $39.96 \pm 0.34$ & $32.73 \pm 0.30$ & $71.36 \pm 0.94$ \\
 & CLIP\cite{radford2021clip}         & - & $43.10 \pm 0.40$ & $35.81 \pm 0.39$ & $70.41 \pm 2.25$ \\
 & DINO-v3\cite{simeoni2025dinov3}       & - & $46.55 \pm 0.67$ & $39.06 \pm 0.62$ & $63.82 \pm 4.61$ \\
 & Slot-TTA\cite{prabhudesai2023test}     & - & $29.67 \pm 0.51$ & $25.11 \pm 0.44$ & $61.85 \pm 3.67$ \\
 & \textbf{DenseTRF (ours)} & - & $\mathbf{49.03 \pm 0.43}$ & $\mathbf{41.34 \pm 0.27}$ & $\mathbf{53.60 \pm 1.82}$ \\
\hline
\multirow{6}{*}{Cholec}
 & SAM\cite{kirillov2023sam}          & - & $43.36 \pm 1.55$ & $34.90 \pm 1.14$ & $116.19 \pm 1.70$ \\
 & DINO-v1\cite{caron2021dino}      & - & $46.67 \pm 2.21$ & $37.18 \pm 1.82$ & $114.11 \pm 0.90$ \\
 & CLIP\cite{radford2021clip}         & - & $54.28 \pm 2.13$ & $44.00 \pm 2.00$ & $108.02 \pm 4.08$ \\
 & DINO-v3\cite{simeoni2025dinov3}       & - & $55.49 \pm 1.98$ & $46.39 \pm 2.03$ & $94.24 \pm 3.07$ \\
 & Slot-TTA\cite{prabhudesai2023test}     & - & $37.40 \pm 0.81$ & $30.53 \pm 0.78$ & $88.92 \pm 2.54$ \\
 & \textbf{DenseTRF (ours)} & - & $\mathbf{57.83 \pm 1.97}$ & $\mathbf{47.29 \pm 1.83}$ & $\mathbf{69.66 \pm 2.61}$ \\
\hline
\end{tabular}
}
\end{table}
Following the comparison with state-of-the-art segmentation architectures, we further evaluate DenseTRF against representative foundation models, including SAM \cite{kirillov2023sam}, DINO-v1 \cite{caron2021dino}, DINO-v3 \cite{simeoni2025dinov3} , and CLIP \cite{radford2021clip}, as well as the slot-based adaptation baseline (Slot-TTA) \cite{prabhudesai2023test}. To ensure a fair comparison, all foundation models are equipped with the same lightweight MLP dense prediction head used in our method, and training is conducted under an extreme low-data regime using only 1\% of labeled samples. Each experiment is repeated across five independent runs.

As shown in table \ref{tab_FM}, across all datasets, DenseTRF consistently achieves the best performance. On the Thoracic dataset, DenseTRF improves DICE to 79.32\%, surpassing the strongest foundation baseline (CLIP/DINO-v3) by approximately 1.6–1.7 points while also yielding the lowest boundary error (HD). Similar trends are observed on POEM, where DenseTRF reaches 49.03\% DICE and reduces HD by over 10 pixels compared to DINO-v3, demonstrating robust adaptation under severe data scarcity. On the Cholec dataset, DenseTRF again delivers the highest DICE (57.83\%, $p<0.05$) and substantially improves boundary quality, reducing HD by more than 24 pixels relative to DINO-v3. Slot‑TTA tends to struggle across all domains because it relies on a high‑quality slot decoder; although its mask‑classification and remixing approach performs well on synthetic images, it does not transfer effectively to the variability of real‑world images.

These results indicate that simply attaching a segmentation head to strong foundation representations is insufficient for reliable dense prediction under distribution shift. In contrast, DenseTRF effectively refits texture-aware grouping to the target domain, yielding consistent gains in both region overlap and boundary accuracy across diverse surgical procedures.
\subsection{Ablation study}

\begin{table}[t!]
\caption{Ablation study across different datasets.}
\label{tab_abla}
\centering
\setlength\arrayrulewidth{0.9pt}
\setlength\doublerulesep{0.9pt} 
\setlength{\tabcolsep}{4pt} 
\resizebox{1.0\linewidth}{!}{%
\scriptsize 
\begin{tabular}{l ccc ccc}
\toprule
 & \multicolumn{2}{c}{Thoracic} & \multicolumn{2}{c}{POEM} & \multicolumn{2}{c}{Cholec} \\
\cmidrule(lr){2-3} \cmidrule(lr){4-5} \cmidrule(lr){6-7}
Method & DICE $\uparrow$ & HD $\downarrow$ & DICE $\uparrow$ & HD $\downarrow$ & DICE $\uparrow$ & HD$\downarrow$ \\
\midrule
w/o SA            & $77.86 \pm 0.87$ & $57.41 \pm 1.22$ & $44.95 \pm 0.40$ & $64.86 \pm 3.40$ & $56.94 \pm 1.55$ & $95.71 \pm 2.09$ \\
w SA w/o ada.     & $76.34 \pm 0.87$ & $60.43 \pm 1.10$ & $42.76 \pm 0.16$ & $63.08 \pm 2.93$ & $53.33 \pm 0.93$ & $101.30 \pm 1.71$ \\
w/o concat        & $77.86 \pm 0.09$ & $60.88 \pm 0.32$ & $42.57 \pm 0.26$ & $66.76 \pm 1.23$ & $55.07 \pm 0.74$ & $102.72 \pm 1.06$ \\
\textbf{Our full} & $\mathbf{79.32} \pm \mathbf{0.48}$ & $\mathbf{55.35} \pm \mathbf{0.09}$ & $\mathbf{49.03} \pm \mathbf{0.43}$ & $\mathbf{53.60} \pm \mathbf{1.82}$ & $\mathbf{57.83} \pm \mathbf{1.97}$ & $\mathbf{69.66} \pm \mathbf{2.61}$ \\
\bottomrule
\end{tabular}
}
\end{table}
We conduct ablation studies to analyze the contribution of each component in DenseTRF, including SA representation learning, test-distribution adaptation, and multi-source representation concatenation. All variants share the same backbone and training protocol to ensure a controlled comparison.

\noindent\textbf{w/o SA}: Our full architecture trained without the SA reconstruction loss and without initializing from a pretrained SA. This baseline underperforms the full model on all datasets, with notably lower Dice on POEM and Cholec, and higher HD on Thoracic and Cholec, e.g., Thoracic DICE drops from 79.32\% to 77.86\% and Cholec DICE from 57.83\% to 56.94\%.

\noindent\textbf{w SA w/o ada}: The model is initialized with a pretrained SA model but is not adapted to the test distribution during training. Interestingly, this often leads to worse performance compared to w/o SA, especially on Thoracic (DICE drops from 77.86 to 76.34) and POEM (DICE 42.76 vs. 44.95). This indicates that simply incorporating a pretrained SA without task-specific adaptation can be detrimental, likely due to a domain shift. 

\noindent\textbf{w/o concat}: This variant uses the same training and initialization strategy as the full model, but instead of concatenating the adapted feature with the SA mask and reconstructed feature, it relies solely on the adapted feature. The results show a clear drop in performance compared to the full model across all metrics, particularly in boundary quality (Cholec HD: 102.72 vs 69.66, $p <0.01$), highlighting the importance of fusing multi-modal cues (feature, mask, reconstruction) for accurate segmentation.

These ablations show that DenseTRF’s gains arise from the combination of slot grouping, target-domain adaptation, and feature fusion. Pretrained slots alone are insufficient, while the full model improves both overlap and boundary accuracy, highlighting the value of adapted slot representations for surgical dense prediction.

\section{Conclusion}

We presented DenseTRF, a texture-aware test-distribution adaptation fra\-mework for surgical dense prediction that combines slot-based object-centric re\-pre\-sen\-tations with unsupervised periodic model merging. By specializing texture-centric slots to the target domain without annotations and fusing encoder features, reconstructions, and attention masks, DenseTRF effectively mitigates distribution shift. Experiments on Thoracic, POEM, and Cholec datasets demonstrate consistent improvements over state-of-the-art segmentation and foundation model baselines under extreme low-data regimes. Ablation results further validate the contribution of each component. Overall, DenseTRF high\-lights object-\-centric learning as a promising strategy for robust domain-adaptive surgical scene understanding.
 
\subsubsection{Acknowledgements.} 
This work was supported by the Linda Pechenik Montague Investigator Award, the American Surgical Association Foundation Fellowship, and Penn AI fellowship. We also acknowledge Kenta Nakahashi and Surgical AI Research Academy of University Health Network for sharing the thoracic dataset.

\subsubsection{Disclosure of interests.} The authors have no competing interests to declare.
%
%
%
\bibliography{refs_miccai26.bib}

@article{janowczyk2016deep,
 author = {Janowczyk, Andrew and Madabhushi, Anant},
 journal = {Journal of pathology informatics},
 number = {1},
 pages = {29},
 title = {Deep learning for digital pathology image analysis: A comprehensive tutorial with selected use cases},
 volume = {7},
 year = {2016}
}

@article{madani2022artificial,
 author = {Madani, Amin and Namazi, Babak and Altieri, Maria S and Hashimoto, Daniel A and Rivera, Angela Maria and Pucher, Philip H and Navarrete-Welton, Allison and Sankaranarayanan, Ganesh and Brunt, L Michael and Okrainec, Allan and others},
 journal = {Annals of surgery},
 number = {2},
 pages = {363--369},
 title = {Artificial intelligence for intraoperative guidance: using semantic segmentation to identify surgical anatomy during laparoscopic cholecystectomy},
 volume = {276},
 year = {2022}
}

@article{hashimoto2018artificial,
 author = {Hashimoto, Daniel A and Rosman, Guy and Rus, Daniela and Meireles, Ozanan R},
 journal = {Annals of surgery},
 number = {1},
 pages = {70--76},
 title = {Artificial intelligence in surgery: promises and perils},
 volume = {268},
 year = {2018}
}

@article{xie2022unsupervised,
 author = {Xie, Qingsong and Li, Yuexiang and He, Nanjun and Ning, Munan and Ma, Kai and Wang, Guoxing and Lian, Yong and Zheng, Yefeng},
 journal = {IEEE Transactions on Medical Imaging},
 number = {1},
 pages = {4--14},
 title = {Unsupervised domain adaptation for medical image segmentation by disentanglement learning and self-training},
 volume = {43},
 year = {2024}
}

@inproceedings{Wang_2022_CVPR,
 author = {Qin Wang and
Olga Fink and
Luc Van Gool and
Dengxin Dai},
 booktitle = {{IEEE/CVF} Conference on Computer Vision and Pattern Recognition,
{CVPR} 2022, New Orleans, LA, USA, June 18-24, 2022},
 pages = {7191--7201},
 title = {Continual Test-Time Domain Adaptation},
 year = {2022}
}

@inproceedings{wang2021tent,
 author = {Dequan Wang and
Evan Shelhamer and
Shaoteng Liu and
Bruno A. Olshausen and
Trevor Darrell},
 booktitle = {9th International Conference on Learning Representations, {ICLR} 2021,
Virtual Event, Austria, May 3-7, 2021},
 title = {Tent: Fully Test-Time Adaptation by Entropy Minimization},
 year = {2021}
}

@inproceedings{locatello2020object,
 author = {Francesco Locatello and
Dirk Weissenborn and
Thomas Unterthiner and
Aravindh Mahendran and
Georg Heigold and
Jakob Uszkoreit and
Alexey Dosovitskiy and
Thomas Kipf},
 booktitle = {Advances in Neural Information Processing Systems 33: Annual Conference
on Neural Information Processing Systems 2020, NeurIPS 2020, December
6-12, 2020, virtual},
 title = {Object-Centric Learning with Slot Attention},
 year = {2020}
}

@inproceedings{seitzer2022bridging,
 author = {Maximilian Seitzer and
Max Horn and
Andrii Zadaianchuk and
Dominik Zietlow and
Tianjun Xiao and
Carl{-}Johann Simon{-}Gabriel and
Tong He and
Zheng Zhang and
Bernhard Sch{\"{o}}lkopf and
Thomas Brox and
Francesco Locatello},
 booktitle = {The Eleventh International Conference on Learning Representations,
{ICLR} 2023, Kigali, Rwanda, May 1-5, 2023},
 title = {Bridging the Gap to Real-World Object-Centric Learning},
 year = {2023}
}

@article{maier2024metrics,
 author = {Maier-Hein, Lena and Reinke, Annika and Godau, Patrick and Tizabi, Minu D and Buettner, Florian and Christodoulou, Evangelia and Glocker, Ben and Isensee, Fabian and Kleesiek, Jens and Kozubek, Michal and others},
 journal = {Nature methods},
 number = {2},
 pages = {195--212},
 title = {Metrics reloaded: recommendations for image analysis validation},
 volume = {21},
 year = {2024}
}

@inproceedings{nguyen2024adapting,
 author = {Nguyen, Minh and Wang, Alan Q and Kim, Heejong and Sabuncu, Mert R},
 booktitle = {European Conference on Computer Vision},
 organization = {Springer},
 pages = {230--246},
 title = {Adapting to shifting correlations with unlabeled data calibration},
 year = {2024}
}

@inproceedings{prabhudesai2023test,
 author = {Mihir Prabhudesai and
Anirudh Goyal and
Sujoy Paul and
Sjoerd van Steenkiste and
Mehdi S. M. Sajjadi and
Gaurav Aggarwal and
Thomas Kipf and
Deepak Pathak and
Katerina Fragkiadaki},
 booktitle = {International Conference on Machine Learning, {ICML} 2023, 23-29 July
2023, Honolulu, Hawaii, {USA}},
 pages = {28151--28166},
 title = {Test-time Adaptation with Slot-Centric Models},
 volume = {202},
 year = {2023}
}

@inproceedings{radford2021clip,
 author = {Alec Radford and
Jong Wook Kim and
Chris Hallacy and
Aditya Ramesh and
Gabriel Goh and
Sandhini Agarwal and
Girish Sastry and
Amanda Askell and
Pamela Mishkin and
Jack Clark and
Gretchen Krueger and
Ilya Sutskever},
 booktitle = {Proceedings of the 38th International Conference on Machine Learning,
{ICML} 2021, 18-24 July 2021, Virtual Event},
 pages = {8748--8763},
 title = {Learning Transferable Visual Models From Natural Language Supervision},
 volume = {139},
 year = {2021}
}

@inproceedings{caron2021dino,
 author = {Mathilde Caron and
Hugo Touvron and
Ishan Misra and
Herv{\'{e}} J{\'{e}}gou and
Julien Mairal and
Piotr Bojanowski and
Armand Joulin},
 booktitle = {2021 {IEEE/CVF} International Conference on Computer Vision, {ICCV}
2021, Montreal, QC, Canada, October 10-17, 2021},
 pages = {9630--9640},
 title = {Emerging Properties in Self-Supervised Vision Transformers},
 year = {2021}
}

@inproceedings{kirillov2023sam,
 author = {Alexander Kirillov and
Eric Mintun and
Nikhila Ravi and
Hanzi Mao and
Chlo{\'{e}} Rolland and
Laura Gustafson and
Tete Xiao and
Spencer Whitehead and
Alexander C. Berg and
Wan{-}Yen Lo and
Piotr Doll{\'{a}}r and
Ross B. Girshick},
 booktitle = {{IEEE/CVF} International Conference on Computer Vision, {ICCV} 2023,
Paris, France, October 1-6, 2023},
 pages = {3992--4003},
 title = {Segment Anything},
 year = {2023}
}

@inproceedings{xie2021segformer,
 author = {Enze Xie and
Wenhai Wang and
Zhiding Yu and
Anima Anandkumar and
Jose M. Alvarez and
Ping Luo},
 booktitle = {Advances in Neural Information Processing Systems 34: Annual Conference
on Neural Information Processing Systems 2021, NeurIPS 2021, December
6-14, 2021, virtual},
 pages = {12077--12090},
 title = {SegFormer: Simple and Efficient Design for Semantic Segmentation with
Transformers},
 year = {2021}
}

@inproceedings{he2016resnet,
 author = {Kaiming He and
Xiangyu Zhang and
Shaoqing Ren and
Jian Sun},
 booktitle = {2016 {IEEE} Conference on Computer Vision and Pattern Recognition,
{CVPR} 2016, Las Vegas, NV, USA, June 27-30, 2016},
 pages = {770--778},
 title = {Deep Residual Learning for Image Recognition},
 year = {2016}
}

@inproceedings{ronneberger2015unet,
 author = {Ronneberger, Olaf and Fischer, Philipp and Brox, Thomas},
 booktitle = {MICCAI},
 title = {U-Net: Convolutional Networks for Biomedical Image Segmentation},
 year = {2015}
}

@inproceedings{hatamizadeh2022swinunetr,
 author = {Hatamizadeh, Ali and others},
 booktitle = {MICCAI},
 title = {Swin UNETR: Swin Transformers for semantic segmentation of brain tumors in MRI images},
 year = {2022}
}

@inproceedings{gao2024segman,
 author = {Fu, Yunxiang and Lou, Meng and Yu, Yizhou},
 booktitle = {CVPR},
 title = {SegMAN: Omni-scale contextual modeling for semantic segmentation},
 year = {2024}
}

@inproceedings{cheng2022mask2former,
 author = {Bowen Cheng and
Ishan Misra and
Alexander G. Schwing and
Alexander Kirillov and
Rohit Girdhar},
 booktitle = {{IEEE/CVF} Conference on Computer Vision and Pattern Recognition,
{CVPR} 2022, New Orleans, LA, USA, June 18-24, 2022},
 pages = {1280--1289},
 title = {Masked-attention Mask Transformer for Universal Image Segmentation},
 year = {2022}
}

@inproceedings{kakogeorgiou2024spot,
 author = {Ioannis Kakogeorgiou and
Spyros Gidaris and
Konstantinos Karantzalos and
Nikos Komodakis},
 booktitle = {{IEEE/CVF} Conference on Computer Vision and Pattern Recognition,
{CVPR} 2024, Seattle, WA, USA, June 16-22, 2024},
 pages = {22776--22786},
 title = {{SPOT:} Self-Training with Patch-Order Permutation for Object-Centric
Learning with Autoregressive Transformers},
 year = {2024}
}

@inproceedings{yang2025continual,
 author = {Yang, Enneng and Tang, Anke and Shen, Li and Guo, Guibing and Wang, Xingwei and Cao, Xiaochun and Zhang, Jie},
 booktitle = {The Thirty-ninth Annual Conference on Neural Information Processing Systems},
 title = {Continual Model Merging without Data: Dual Projections for Balancing Stability and Plasticity},
 year = {2025}
}

@article{simeoni2025dinov3,
 author = {Sim{\'e}oni, Oriane and Vo, Huy V and Seitzer, Maximilian and Baldassarre, Federico and Oquab, Maxime and Jose, Cijo and Khalidov, Vasil and Szafraniec, Marc and Yi, Seungeun and Ramamonjisoa, Micha{\"e}l and others},
 journal = {ArXiv preprint},
 title = {Dinov3},
 volume = {abs/2508.10104},
 year = {2025}
}

@inproceedings{liao2025forla,
 author = {Liao, Guiqiu and Jogan, Matjaz and Eaton, Eric and Hashimoto, Daniel A},
 booktitle = {The Thirty-ninth Annual Conference on Neural Information Processing Systems},
 title = {FORLA: Federated Object-Centric Representation Learning with Slot Attention},
 year = {2025}
}

@inproceedings{didolkar2024zero,
 author = {Didolkar, Aniket and Zadaianchuk, Andrii and Goyal, Anirudh and Mozer, Mike and Bengio, Yoshua and Martius, Georg and Seitzer, Maximilian},
 booktitle = {The Thirteenth International Conference on Learning Representations},
 title = {On the Transfer of Object-Centric Representation Learning},
 year = {2025}
}

@article{twinanda2016endonet,
 author = {Twinanda, Andru P and Shehata, Sherif and Mutter, Didier and Marescaux, Jacques and De Mathelin, Michel and Padoy, Nicolas},
 journal = {IEEE transactions on medical imaging},
 number = {1},
 pages = {86--97},
 title = {Endonet: a deep architecture for recognition tasks on laparoscopic videos},
 volume = {36},
 year = {2016}
}

@article{hong2020cholecseg8k,
 author = {Hong, W-Y and Kao, C-L and Kuo, Y-H and Wang, J-R and Chang, W-L and Shih, C-S},
 journal = {ArXiv preprint},
 title = {Cholecseg8k: a semantic segmentation dataset for laparoscopic cholecystectomy based on cholec80},
 volume = {abs/2012.12453},
 year = {2020}
}
\bibliographystyle{splncs04}
 
\end{document}